\documentclass[letterpaper,10pt,conference]{ieeeconf}
\IEEEoverridecommandlockouts
\usepackage[T1]{fontenc}
\usepackage{times}
\usepackage{amsmath,amssymb}
\usepackage{graphicx}
\usepackage{booktabs}
\usepackage{array}
\usepackage{xcolor}
\usepackage{url}
\graphicspath{{figures/}}
\definecolor{draftorange}{RGB}{157,76,16}
\newcommand{\method}{AnyviewMeter}

\title{AnyviewMeter: Adapting Robotic Reward Models\\with Camera Geometry and Multi-View Attention}
\author{Yuang Tu, Runjia Tan, Yujie Yan, Jinghan Hu, and Chen Lv$^{*}$\\[2pt]
Nanyang Technological University, Singapore%
\thanks{$^{*}$Corresponding author: Chen Lv (lyuchen@ntu.edu.sg).}}
\begin{document}
\maketitle
\thispagestyle{empty}\pagestyle{empty}

\begin{abstract}
Robotic reward models evaluate task execution from visual observations, but their predictions can change with camera viewpoint and occlusion even when the underlying task state is unchanged. Adapting a pretrained reward model to a local task therefore requires accounting for how that task is observed. We introduce \method, a geometry-conditioned adaptation framework for robotic reward models that represent task progress as a scalar reward signal. It combines low-rank fine-tuning with token-aligned Pl\"ucker rays and synchronous block attention: ray conditioning incorporates camera geometry into visual features and attention queries and keys, while block attention fuses synchronized views inside the pretrained decoder. The framework supports both single-view reward prediction and joint multi-view evaluation through parameter-efficient adaptation of a pretrained Robometer model. On PickCube, single-view adaptation improves progress prediction in every camera group and reduces mean absolute error under a changed field of view by approximately 21\% relative to RGB fine-tuning. Across simulated manipulation tasks, joint multi-view prediction reduces progress error by 41--69\% compared with averaging single-view RGB predictions and improves temporal ordering in approximately 88\% of task--camera groups. On real tasks with fixed and wrist-mounted cameras, mean absolute error decreases by approximately 21\% relative to averaged RGB fine-tuning. These results support camera geometry and joint visual evidence as useful components of task-specific robotic reward adaptation.
\end{abstract}

\section{Introduction}
Robotic reward models turn visual observations into signals for evaluating task execution. When a model represents reward through task progress, its predictions should reflect the task state across different camera observations. A camera ray connects each image location to a viewing direction in the physical world, making observation geometry relevant to reward prediction. A gripper approaching a target may move visibly in one camera and almost entirely along the viewing direction in another. A cube resting on another cube and a cube positioned behind it may have similar projections, despite representing different task outcomes. Progress is a property of the task state, but the evidence available to estimate it depends on the camera.

General-purpose reward models such as Robometer~\cite{robometer} make it possible to begin with a pretrained understanding of manipulation rather than build a task evaluator from scratch. A local deployment nevertheless has its own task, workspace, and camera arrangement. The practical question is how to adapt that existing model to these conditions with a limited set of task trajectories and a manageable training footprint. Ordinary RGB fine-tuning can improve task fit, but it leaves the model to infer the observation geometry from appearance alone.

We study \emph{Pl\"ucker-conditioned adaptation of robotic reward models}. A calibrated camera defines a six-dimensional ray at each visual token. We use these rays as an additional input while adapting a pretrained reward model through LoRA~\cite{lora}, its progress head, and a geometry module. This formulation first addresses the single-camera setting: the model receives both visual evidence and the geometry under which that evidence was obtained. It then extends to synchronized cameras, where multiple observations of the same state share an attention sequence. Geometry enters both the visual features and the queries and keys used to combine them.

\begin{figure*}[t]
\centering\includegraphics[width=.97\textwidth]{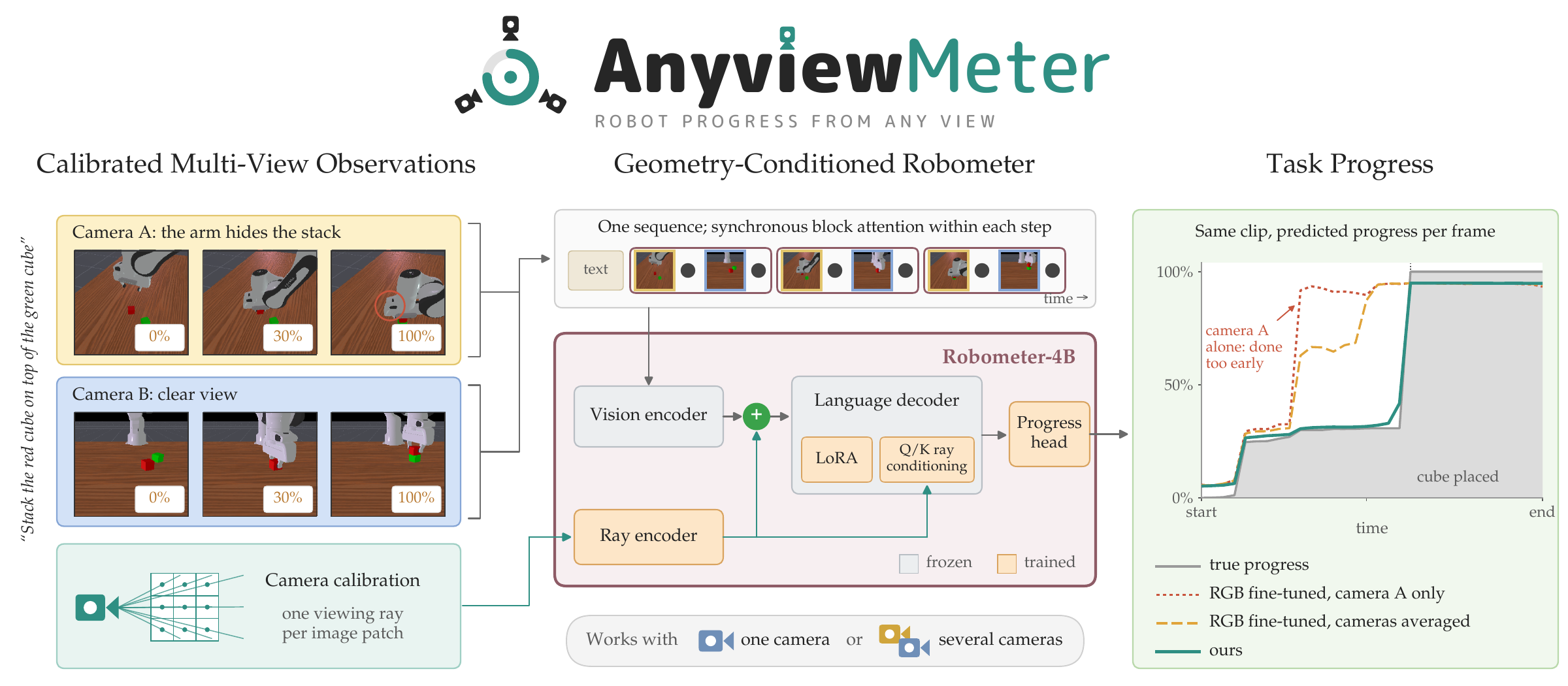}
\caption{Overview of \method. Left: one StackCube test execution seen by a camera whose view of the stack is blocked by the arm (A) and by a clear camera (B); tags give the state-derived progress label. Calibration assigns a Pl\"ucker ray to each image patch. Middle: synchronized frames from both cameras form one sequence for the Robometer-4B backbone, in which synchronous block attention lets the two frames of each time step (outlined) attend to each other. A ray encoder adds geometry to visual tokens and, through Q/K ray conditioning, to decoder queries and keys; base weights stay frozen while LoRA, the geometry modules, and the progress head are trained. The same model accepts one or several cameras. Right: measured per-frame predictions on this clip.}
\label{fig:method}
\end{figure*}

Camera geometry helps interpret an observation but cannot recover evidence hidden by occlusion; a complementary camera can. We therefore fuse complementary views inside the decoder, rather than averaging per-view predictions, using synchronous block attention, which permits mutual attention within each instant while preserving temporal causality. We evaluate single-view generalization and joint multi-view prediction against late fusion in simulation and on a real robot.

We adapt a separate checkpoint for each task from the same pretrained model. Our contributions are (i) a geometry-conditioned framework for adapting robotic reward models to task-specific progress prediction from one or several cameras; (ii) synchronous block attention, an attention mask that fuses synchronized views inside the pretrained decoder without new parameters; and (iii) an evaluation in simulation and on a real robot with a fixed and a wrist-mounted camera, showing that single-view adaptation generalizes better to unseen viewpoints and that joint multi-view prediction outperforms late fusion.

\section{Related Work}
Prior work connects visual task evaluation with representation learning, multi-view perception, and adaptation of pretrained models. We organize these connections around the information available to the evaluator and how it is incorporated.

\paragraph{Visual Rewards and Progress Estimation}
Visual reward learning uses goal-conditioned representations or vision-language feedback to evaluate task completion from observations~\cite{vip,liv,rlvlmf}. More recent robot evaluators learn progress, success, progress changes, or temporal distance from heterogeneous experience~\cite{robometer,gvl,roboreward,vlac,rynnvalue,robodopamine}. Their supervision includes demonstrations, trajectory preferences, relabeled outcomes, and temporal targets. Among these, Robometer~\cite{robometer} combines frame-level progress with trajectory comparisons, while RynnValue~\cite{rynnvalue} predicts remaining time and suppresses temporal shortcuts. These objectives support different reward interfaces; scalar calibration and temporal ordering must therefore be evaluated separately. We study how calibrated camera observations improve task adaptation of an existing progress evaluator.

\paragraph{Multi-View Representations and Fusion}
Temporal correspondence across synchronized views or different executions provides supervision for task-relevant visual representations~\cite{tcn,tcc}. Joint attention also combines complementary inputs in scene representation, multi-scale recognition, multi-camera perception, and robot manipulation~\cite{srt,crossvit,petr,octo,mvmwm}. Robo-Dopamine likewise learns a process reward model from multi-view inputs~\cite{robodopamine}. These settings motivate shared processing, but agreement between views alone does not establish that either view observes the evidence needed for progress estimation. Unlike these, we fuse synchronized cameras inside a pretrained reward model to expose state changes obscured in one projection.

\paragraph{Geometric Conditioning}
Classical multi-view geometry relates projections and combines geometric constraints with appearance for correspondence estimation~\cite{multiviewgeometry,linegeometry}. Learned models likewise incorporate camera or spatial information into visual features and attention for detection, rendering, pose estimation, and generation~\cite{srt,petr,lfn,raydiffusion,gta,prope,cameractrl,scope}. CameraCtrl~\cite{cameractrl} uses per-pixel Pl\"ucker embeddings for camera control, and SCoPE~\cite{scope} introduces sightline-dependent query/key terms while preserving pretrained positional encoding. In robot learning, vision-language-action policies are sensitive to camera viewpoint~\cite{liberoplus}; conditioning policies on per-pixel Pl\"ucker rays improves viewpoint generalization~\cite{doyouknow}, and cross-view consistency objectives require no camera inputs at inference~\cite{crossview}. We bring feature-level and attention-level ray conditioning to progress estimation.

\paragraph{Temporal Attention and Readout Isolation}
Attention masks can separately control temporal visibility and feedback from prediction tokens~\cite{rynnvalue,octo}. Octo permits time-block causal observation attention while preventing observations and task tokens from reading readouts. RynnValue additionally isolates value-query groups across observations and prevents context tokens from relaying their information. Our synchronous block attention instead groups the synchronized views of each instant and isolates only each frame's own readout (Sec.~III-D).

\section{Pl\"ucker-Conditioned Task Adaptation}
\subsection{Problem Formulation}
For a task instruction $\ell$, let $I_t^v$ be the image at time $t$ from camera $v$, with intrinsic matrix $K_v$ and world-to-camera extrinsics $[R_v\mid t_v]$, which change over time for a wrist-mounted camera. The model predicts normalized progress $\hat y_t\in[0,1]$ from an observed sequence and its camera parameters. Training uses a task-specific dataset with state-derived progress labels $y_t$ where available. All views of a physical execution belong to the same dataset split.

Let $\theta_0$ denote the pretrained backbone, $\Delta\theta$ its LoRA updates~\cite{lora}, $\psi$ the geometry parameters, and $\omega$ the progress head. We optimize $(\Delta\theta,\psi,\omega)$ while keeping $\theta_0$ fixed. Following the discrete progress interface of Robometer~\cite{robometer}, the head predicts ten probabilities $p_{t,b}$, with centers $c_b=(b+1/2)/10$ for $b\in\{0,\ldots,9\}$:
\begin{equation}
  \hat y_t=\sum_{b=0}^{9} c_b p_{t,b}.
  \label{eq:progress}
\end{equation}
The same prediction interface is used for one or two cameras. A downstream online evaluator must be queried only on the available history; offline full-clip metrics alone do not establish causal online performance.

\subsection{Token-Aligned Camera Rays}
Using calibrated back-projection and the direction--moment representation of a line~\cite{multiviewgeometry,cameractrl,pottmann}, for a token centered at image coordinate $(u,v)$, the world-frame camera center, ray direction, and moment are
\begin{align}
  c &= -R^\top t, &
  d &= \frac{R^\top K^{-1}[u,v,1]^\top}
  {\|R^\top K^{-1}[u,v,1]^\top\|_2},\\
  m &= c\times d, & r&=(d,m).\label{eq:ray}
\end{align}
We sample these rays at token centers, after accounting for the visual encoder's spatial merging. At $256\times256$ resolution, patch size 16 and a $2\times2$ merge yield an $8\times8$ grid. Camera parameters and token order must describe the same image transformation.

The moment is kept in metric coordinates rather than normalized independently of the direction. For two raw rays, the reciprocal product~\cite{pottmann}
\begin{equation}
  \mathcal R(r_i,r_j)=d_i^\top m_j+m_i^\top d_j
  \label{eq:reciprocal}
\end{equation}
vanishes exactly when the two lines are coplanar, that is, when they intersect or are parallel. Rays through a common camera center therefore always give zero. For rays with different centers, whether from two cameras or from one moving camera, the reciprocal product is zero when their supporting lines intersect or are parallel. A nonzero value therefore rules out a common observed point; zero alone does not establish one. The term does not by itself estimate depth or visibility.

\subsection{Feature and Attention Conditioning}
\paragraph{Feature-level adaptation}
A ray encoder $g_\psi$ maps $r_i$ to the visual hidden dimension and adds it to the corresponding merged visual token:
\begin{equation}
 x_i'=x_i+g_\psi(r_i).
\end{equation}
The output projection is initialized to zero, preserving the pretrained visual features at initialization, following the zero-initialized conditioning principle used in ControlNet~\cite{controlnet}.

\paragraph{Query/key adaptation}
Following the sightline-coordinate formulation~\cite{scope} and related camera-aware attention~\cite{gta,prope}, \method{} also applies ray-dependent updates after the pretrained positional encoding:
\begin{equation}
  q_i'=q_i+\alpha_\ell a_i,\qquad
  k_j'=k_j+\alpha_\ell b_j.
\end{equation}
Here $a_i$ and $b_j$ are projected ray features. Direction and moment are swapped in the key representation, and an input-dependent gate uses $\log(\|m\|+\epsilon)$. Expanding the attention numerator gives
\begin{align}
(q_i')^\top k_j'={}&q_i^\top k_j
 +\alpha_\ell(q_i^\top b_j+a_i^\top k_j)\nonumber\\
 &+\alpha_\ell^2 a_i^\top b_j.\label{eq:attention}
\end{align}
The terms represent appearance, appearance--geometry interactions, and a ray-only interaction. A geometric projection initialization makes the last term proportional to Eq.~\eqref{eq:reciprocal} under the initial uniform gate, while $\alpha_\ell=0$ initially preserves pretrained attention. Learned projections and gates need not retain the exact reciprocal-product structure. The default \method{} combines both paths, with query/key updates at decoder layers 27, 31, and 35; our single-view experiments use the patch path alone.

\subsection{Joint Multi-View Prediction}
For synchronized cameras, we interleave frames as
\begin{equation}
 (I_1^1,I_1^2,I_2^1,I_2^2,\ldots,I_T^1,I_T^2).
\end{equation}
Camera rays follow this same ordering. The frames share the backbone attention sequence.

\begin{figure}[t]
\centering\includegraphics[width=\columnwidth]{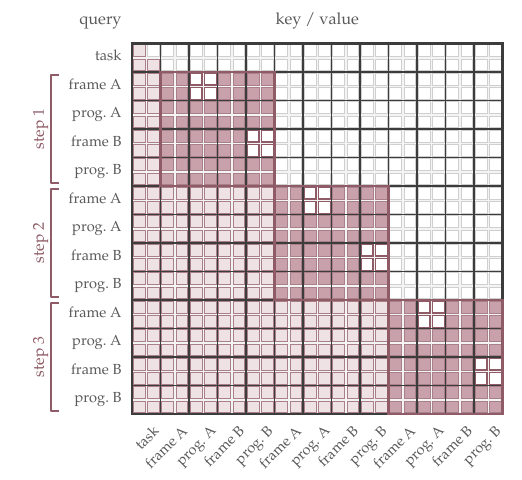}
\caption{Synchronous block attention for two cameras. Frame tokens of cameras A and B are interleaved, each followed by its progress token; rows are queries and columns keys/values. Each cell is a token group drawn as $2\times2$ schematic tokens: dark squares are visible within the same time step, light squares are visible earlier tokens, and white squares are masked. The task text stays token-causal. Within a time step all groups attend to each other, except that a frame cannot attend to its own progress token; later time steps are masked.}
\label{fig:blockattn}
\end{figure}

\paragraph{Synchronous block attention}
Robometer's Qwen3-VL decoder~\cite{robometer,qwen3vl} uses a token-causal mask, which suits a single camera: each progress token reads its own frame and the past. With interleaved cameras, however, the first camera's tokens cannot attend to the second camera's frame at the same instant, whereas the second camera's tokens can attend to the first, so fusion is one-directional and depends on the camera order. Synchronous block attention makes all views of an instant attend to each other while no token attends to a future instant (Fig.~\ref{fig:blockattn}). Let $\mathcal B_t$ span the tokens from the first view's visual-start token through the last view's progress token at time $t$, including intervening tokens; let $\mathcal F_t^v\subset\mathcal B_t$ be the tokens of view $v$'s frame, from its visual-start to its visual-end token, and $p_t^v$ its progress token. For query position $i$ and key position $j$, the additive attention mask takes the first matching case:
\[
 M_{ij}=\begin{cases}
 -\infty, & \exists t,v:\ i\in\mathcal F_t^v,\ j=p_t^v,\\
 0, & j\leq i\ \text{or}\ \exists t:\ i,j\in\mathcal B_t,\\
 -\infty, & \text{otherwise}.
 \end{cases}
\]
The first case is a self-readout mask. Block visibility would also let a frame read its own progress token, a summary of that same frame. We remove only this path, so the readout a frame receives within an instant comes from the other camera, which favors fusion across cameras (Table~\ref{tab:ablation_attn}). Otherwise each time block is fully visible, earlier blocks remain visible, and future blocks and padding keys are masked; prompt and inter-block tokens keep token-causal attention. Progress tokens are learned readout tokens rather than labels, so no target values are exposed. The mask is applied at the decoder input after M-RoPE position construction and is shared by all layers. It adds no trainable parameters, and because positions remain order-dependent, it does not guarantee invariance to camera order.

\paragraph{Progress readout}
Under synchronous block attention, the progress token of every view at time $t$ attends to all views of that instant. We read the token of the last view, $v=V$:
\begin{equation}
 \hat y_t=\sum_b c_b\,p^{V}_{t,b}.
 \label{eq:readout}
\end{equation}
There is no additional learned fusion head. This differs from averaging independently scored videos, because the visual representations have already been jointly processed. The implementation also supports averaging the per-view progress tokens.

\subsection{Task Supervision and Adaptation Budget}
Categorical prediction of scalar values has been studied as an alternative to direct regression~\cite{bellemare,stopregressing}. Here a continuous target $y_t$ is split linearly between its two nearest bin centers, giving soft label $q_{t,b}$; targets below the first center or above the last, including the endpoints 0 and 1, put all mass on the end bin. With per-frame weight $w_t$ derived from training-bin frequencies, the progress objective is
\begin{align}
 \mathcal L_{\rm prog}&=\frac{1}{|\mathcal T_L|}\sum_{t\in\mathcal T_L} w_t
 \left[\mathcal L_{\rm bin,t}+\tfrac12(\hat y_t-y_t)^2\right],\nonumber\\
 \mathcal L_{\rm bin,t}&=-\sum_b q_{t,b}\log\bar p_{t,b}.
\end{align}
Here $\mathcal T_L$ contains labeled time steps, and $\bar p_{t,b}$ is the distribution that forms $\hat y_t$: the single view's progress token, or the last view's under joint prediction (Eq.~\eqref{eq:readout}); the causal ablations in Table~\ref{tab:ablation_attn} average the views' distributions instead. Because $\hat y_t$ is an expectation over the bin centers, it lies in $[0.05,0.95]$. The code also supports consistency between separate single-view predictions, but the reported joint-view runs use zero consistency weight. Their gains therefore arise without an explicit consistency penalty. Unlabeled failure clips are excluded from supervised progress error and do not acquire a synthetic progress curve.

\method{} trains 74.39M parameters: 66.06M LoRA parameters, a 3.29M progress head, and a 5.04M geometry adapter (patch ray encoder and Q/K ray adapters), about 7.3\% more than RGB adaptation. Two-view training peaks at 13.43G of GPU memory at about 0.46 s per step on one RTX 5090, and a full 24k-step run, including validation and testing, took about 3.6 hours.

\section{Experiments}
We first evaluate single-view generalization and joint multi-view prediction in simulation, then report a real-robot evaluation on three tasks and ablations of joint prediction.

\subsection{Experimental Setup}
\begin{table}[t]
\centering\small
\caption{Implementation settings, shared by all tasks unless noted. Per-task batteries differ in elevation: the seen-azimuth group spans 20--36$^\circ$ on PickCube and 36--43$^\circ$ on PegInsertionSide.}
\label{tab:setup}
\setlength{\tabcolsep}{3pt}
\begin{tabular}{@{}>{\raggedright\arraybackslash}p{.27\columnwidth}>{\raggedright\arraybackslash}p{.69\columnwidth}@{}}\toprule
Setting & Value\\\midrule
Trajectories & 420 train / 60 val / 120 test per task\\
Training labels & 336 labeled, 84 unlabeled\\
Trajectory kinds & 360 success, 120 recovery, 120 failure\\
Training cameras & azimuth $\pm90^\circ$, elevation 5--50$^\circ$\\
Test battery & 17 cameras: seen, edge and far azimuth, changed FOV (four each), one canonical\\
Clips & 32 frames at $256\times256$; 5 sampled per step\\
Backbone & Robometer-4B, base weights frozen\\
LoRA & rank 32, dropout 0.05, all 36 decoder layers, attention and MLP projections\\
Geometry & patch ray encoder; Q/K at layers 27, 31, 35\\
Two-view attention & synchronous block with self-readout mask; last view's progress token read\\
Optimizer & AdamW, weight decay 0.01, gradient clip 1.0\\
Learning rates & $2\times10^{-5}$ LoRA and head; $3\times10^{-4}$ geometry\\
Batch & one clip per step (both views when two-view)\\
Steps & up to 24k\\\bottomrule
\end{tabular}
\end{table}

\begin{table*}[t]
\centering\small
\caption{Progress prediction on four tasks (MAE$\downarrow$ / $\tau_a\uparrow$ from per-clip test predictions); each battery camera is paired with the fixed canonical camera. R0: original Robometer; R1: RGB fine-tuning; P1: \method. ``1 view'' uses the battery camera alone; ``2 views, avg.'' averages a single-view model's predictions on both cameras; P1 processes both views jointly.}
\label{tab:results}
\setlength{\tabcolsep}{5pt}
\begin{tabular}{llcccc}\toprule
Task & Method & Seen azimuth & Edge azimuth & Far azimuth & OOD FOV\\\midrule
PickCube & R0, 1 view & .2024 / .5706 & .2038 / .4764 & .1939 / .5552 & .2159 / .4832\\
 & R0, 2 views, avg. & .1946 / .6419 & .1946 / .5989 & .1902 / .6406 & .2014 / .5985\\
 & R1, 1 view & .1071 / .8228 & .1061 / .7825 & .0989 / .8012 & .1166 / .7913\\
 & R1, 2 views, avg. & .1066 / .8372 & .1053 / .8250 & .1014 / .8313 & .1112 / .8241\\
 & P1 (ours), 2 views & \textbf{.0334} / \textbf{.8797} & \textbf{.0343} / \textbf{.8743} & \textbf{.0335} / \textbf{.8758} & \textbf{.0370} / \textbf{.8774}\\\midrule
PushCube & R0, 1 view & .2423 / .5411 & .2343 / .5911 & .2624 / .5801 & .2447 / .4503\\
 & R0, 2 views, avg. & .2253 / .5598 & .2211 / .5971 & .2344 / .5964 & .2273 / .5010\\
 & R1, 1 view & .0736 / .8151 & .0970 / .7779 & .1633 / .7149 & .0744 / .8190\\
 & R1, 2 views, avg. & .0672 / .8284 & .0772 / .8218 & .1111 / .7822 & .0678 / .8303\\
 & P1 (ours), 2 views & \textbf{.0330} / \textbf{.8630} & \textbf{.0396} / \textbf{.8659} & \textbf{.0430} / \textbf{.8499} & \textbf{.0343} / \textbf{.8636}\\\midrule
StackCube & R0, 1 view & .2841 / .5729 & .2859 / .5522 & .2946 / .5279 & .3016 / .5462\\
 & R0, 2 views, avg. & .2874 / .6199 & .2883 / .6123 & .2927 / .6027 & .2962 / .6095\\
 & R1, 1 view & .1395 / .7110 & .1528 / .7058 & .1352 / .7068 & .1414 / .7223\\
 & R1, 2 views, avg. & .1450 / .7257 & .1516 / .7253 & .1428 / .7221 & .1459 / .7285\\
 & P1 (ours), 2 views & \textbf{.0459} / \textbf{.7720} & \textbf{.0465} / \textbf{.7715} & \textbf{.0451} / \textbf{.7704} & \textbf{.0447} / \textbf{.7709}\\\midrule
PegInsertionSide & R0, 1 view & .1612 / .4961 & .1748 / .5223 & .1549 / .5120 & .1684 / .4381\\
 & R0, 2 views, avg. & .1578 / .5093 & .1506 / .5645 & .1490 / .5425 & .1657 / .4667\\
 & R1, 1 view & .0587 / .7815 & .0893 / .7285 & .1117 / .6933 & .0518 / .7928\\
 & R1, 2 views, avg. & .0515 / .8016 & .0654 / .7784 & .0762 / .7739 & .0485 / .8057\\
 & P1 (ours), 2 views & \textbf{.0280} / \textbf{.8056} & \textbf{.0294} / \textbf{.8016} & \textbf{.0314} / \textbf{.7966} & \textbf{.0288} / \textbf{.8058}\\\bottomrule
\end{tabular}
\end{table*}

\paragraph{Tasks and data}
We use PickCube, PushCube, StackCube, and PegInsertionSide in ManiSkill~\cite{maniskill}. Table~\ref{tab:setup} lists the implementation settings. Each task receives a separate adapted model. The representative 600-trajectory series uses 420 training, 60 validation, and 120 test trajectories. For PickCube, 336 training trajectories have progress targets and 84 are unlabeled under the failure-masking protocol. The test labels used for MAE and temporal ordering are state-derived; unlabeled clips contribute only to label-free prediction-spread measures. Model selection uses validation MAE on held-out camera groups, so those camera families are not completely unseen during selection.

The camera battery separates seen azimuths, edge and far azimuths, and changed field of view (FOV). In joint-view testing, the varying battery camera is paired with a fixed canonical camera. Thus the two-view setting provides a stable auxiliary observation; its gain over one view includes that extra information. All models are trained for up to 24k optimization steps and evaluated at their best validation checkpoint.

\paragraph{Controls and metrics}
R0 is the original Robometer checkpoint; R1 updates LoRA and the progress head using RGB; P1 additionally uses camera geometry. R0 and R1 are single-view models; with two cameras they are run on each camera and their predictions averaged. Two-view P1 processes both cameras jointly under synchronous block attention and reads the last view's progress token; this is the default multi-view configuration of \method{}, and Tables~\ref{tab:ablation_attn} and~\ref{tab:ablation_rays} compare alternatives. We report mean absolute error (MAE), temporal Kendall $\tau_a$, and prediction spread. For the training evaluations, $\tau_a$ normalizes signed pair agreements by all time-step pairs, so tied target plateaus can limit its maximum. Cross-view spread $\sigma_{\rm view}$ is the standard deviation of predictions for the same state, averaged over time and trajectories.

\subsection{Any Single View: Generalization to Unseen Viewpoints}
Single-view \method{} uses the patch path alone (Sec.~III-C). The edge- and far-azimuth test cameras lie 1--61$^\circ$ outside the training azimuth range, and the changed-FOV cameras use narrower 26--32$^\circ$ fields of view; of these, only the changed-FOV group is also excluded from model selection. On PickCube, single-view \method{} improves over single-view RGB adaptation (R1, 1 view in Table~\ref{tab:results}) in every camera group, with MAE 0.078--0.092 versus 0.099--0.117 and higher $\tau_a$ throughout. For changed FOV, MAE decreases from 0.1166 to 0.0918, approximately 21.2\%, while $\tau_a$ increases from 0.7913 to 0.8188. Predictions for the same state also agree more closely across cameras, with cross-view spread 0.034--0.048 versus 0.041--0.057. Both models observe the same single camera and share the 12k budget, so the rays account for the whole difference: they lower MAE by 27.0\%, 14.8\%, 13.9\%, and 21.2\% on the seen-azimuth, edge-azimuth, far-azimuth, and changed-FOV groups. The reduction appears in every group, including the two outside the training azimuth range, but it is not larger there than on the seen azimuths.

\subsection{Any Set of Views: Joint Multi-Camera Input}
Additional views must also be combined appropriately. In a probe of the original Robometer on PickCube (32 frames, 30 trajectories), its best single-camera correlation is 0.788; averaging that camera with the worst reduces it to 0.434, and averaging all five gives 0.501. Visibility-based weighting is not universally reliable either: on PegInsertionSide it gives 0.646, below the 0.743 from equal weighting. These results motivate joint processing.

How the two views are combined matters as much as having them. On PickCube, averaging the predictions of the single-view R1 model on both cameras leaves far-azimuth MAE essentially unchanged (0.0989 to 0.1014) and raises $\tau_a$ from 0.801 to 0.831 (Table~\ref{tab:results}). Late fusion helps more on PushCube and PegInsertionSide, but not on StackCube.

\subsection{Joint Input Versus Late Fusion}
\label{sec:geom}
\method{} has the lowest progress error and the highest ordering score in every camera group of all four tasks (Table~\ref{tab:results}); its MAE is 41--69\% lower than that of late fusion of single-view RGB models. Against late fusion of single-view RGB models, paired trajectory-bootstrap 95\% intervals show lower MAE in every group; $\tau_a$ is higher in every group of PickCube, PushCube, and StackCube and in the edge- and far-azimuth groups of PegInsertionSide, and indistinguishable from late fusion in the remaining two PegInsertionSide groups. The RGB models reach their best validation checkpoints within 8k steps, after which validation error rises, whereas the geometry-conditioned model reaches its best checkpoint between 13k and 19.5k steps.

\subsection{Real-Robot Evaluation}
We evaluate three real tasks recorded with a Franka Panda arm, each observed by a fixed third-person camera and a wrist camera: placing a cup on the plate of the same color (50 demonstrations), picking up a cup (51), and stacking one cup on another (200). Following Robometer's default for demonstrations~\cite{robometer}, progress increases linearly from the first frame and reaches one at 95\% of the episode. R1 is trained on both cameras and averaged over them, and \method{} processes both jointly; the two share the seed, data order, and 2000-step budget, and R0 is untrained.

Each demonstration is subsampled uniformly to 32 frames per camera and resized to $256\times256$ with rescaled intrinsics; camera poses are expressed in the robot base frame, and the wrist camera's recorded pose gives its rays at every frame. Demonstrations are split 70/15/15 into training, validation, and testing. Since time-linear labels can be fit from frame position alone, training clips are reversed (probability 0.5), frozen from a random point at 60--90\% (0.2), or stepped back 4--10 frames at 40--80\% and resumed (0.2), with labels following the displayed frame; four frames per camera are then drawn. Each test demonstration yields a normal, a reversed, a frozen (from 75\%), and a recovery clip (six frames back at 60\%); checkpoints are selected by mean validation MAE over these four clip types.

\begin{figure*}[t]
\centering\includegraphics[width=.97\textwidth]{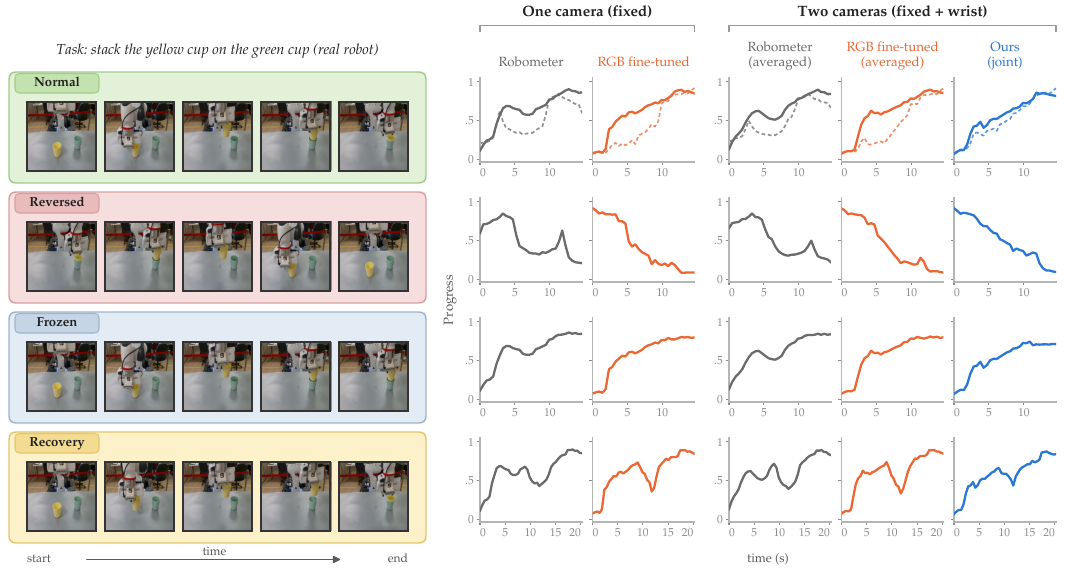}
\caption{Progress predictions on one real stack-cups test demonstration, the median of the 30 test demonstrations by \method{} error, shown as its normal, reversed, frozen, and recovery test clips. Frames are from the fixed camera. With one camera, the original Robometer and RGB fine-tuning see the fixed camera only; with two cameras, both are run on the fixed and wrist cameras and averaged, whereas \method{} processes both jointly. Dashed (top row): each model's prediction on the reversed clip, flipped in time; it should match the solid curve because both clips show the same frames. Time is clip playback time.}
\label{fig:real_curves}
\end{figure*}

\begin{table}[t]
\centering\footnotesize
\caption{Real-robot test MAE$\downarrow$ and $\tau_a\uparrow$, averaged over the four clip types (8, 8, and 30 test demonstrations). R1 is one model scored on each camera alone and on their average.}
\label{tab:real}
\setlength{\tabcolsep}{2.4pt}
\begin{tabular}{lcccccccc}\toprule
& \multicolumn{2}{c}{Cup to plate} & \multicolumn{2}{c}{Pick cup} & \multicolumn{2}{c}{Stack cups} & \multicolumn{2}{c}{Mean}\\
\cmidrule(lr){2-3}\cmidrule(lr){4-5}\cmidrule(lr){6-7}\cmidrule(lr){8-9}
Model & MAE & $\tau_a$ & MAE & $\tau_a$ & MAE & $\tau_a$ & MAE & $\tau_a$\\\midrule
R0, averaged & .154 & .753 & .182 & .592 & .142 & .694 & .160 & .680\\
R1, fixed cam. & .114 & .836 & .090 & .870 & .108 & .840 & .104 & .849\\
R1, wrist cam. & .117 & .838 & .094 & .876 & .110 & .817 & .107 & .844\\
R1, averaged & .114 & .861 & .089 & .896 & .105 & .850 & .103 & .869\\
P1, joint & \textbf{.089} & \textbf{.874} & \textbf{.070} & \textbf{.919} & \textbf{.083} & \textbf{.884} & \textbf{.081} & \textbf{.892}\\\bottomrule
\end{tabular}
\end{table}

\method{} has the lowest MAE and the highest $\tau_a$ on every task in Table~\ref{tab:real}, reducing mean MAE from 0.103 for R1 to 0.081. R1 on either camera alone has mean $\tau_a$ of 0.844--0.849, and averaging the two cameras raises it to 0.869 with little change in MAE. Reversed clips separate the models most, with $\tau_a$ of 0.32--0.53 for R0, 0.85--0.89 for R1, and 0.90--0.93 for \method{}. Fig.~\ref{fig:real_curves} shows the four clips of one test demonstration.

\subsection{Ablation of Joint Prediction}
\begin{table}[t]
\centering\footnotesize
\caption{Attention and readout ablation of joint two-view prediction, far-azimuth MAE$\downarrow$ and $\tau_a\uparrow$, components added one at a time: (a) RGB only, token-causal mask, mean readout, 12k steps; (b) + patch and Q/K rays; (c) causal mask replaced by synchronous block attention, last-view readout; (d) + self-readout mask, i.e., \method{}. (b)--(d) use 24k steps.}
\label{tab:ablation_attn}
\setlength{\tabcolsep}{2.4pt}
\begin{tabular}{lcccccccc}\toprule
& \multicolumn{2}{c}{(a) RGB} & \multicolumn{2}{c}{(b) +rays} & \multicolumn{2}{c}{(c) +block} & \multicolumn{2}{c}{(d) +mask}\\
\cmidrule(lr){2-3}\cmidrule(lr){4-5}\cmidrule(lr){6-7}\cmidrule(lr){8-9}
Task & MAE & $\tau_a$ & MAE & $\tau_a$ & MAE & $\tau_a$ & MAE & $\tau_a$\\\midrule
PickCube & .0494 & .862 & .0416 & \textbf{.887} & .0337 & .863 & \textbf{.0335} & .876\\
PushCube & .0639 & .807 & .0695 & .798 & .0475 & \textbf{.851} & \textbf{.0430} & .850\\
StackCube & .0610 & .752 & .0516 & .755 & .0785 & .729 & \textbf{.0451} & \textbf{.770}\\
PegInsert & .0518 & .765 & .0506 & .771 & .0328 & .787 & \textbf{.0314} & \textbf{.797}\\
\bottomrule
\end{tabular}
\end{table}

Table~\ref{tab:ablation_attn} adds the components of \method{} one at a time to a model that already sees both cameras in one sequence: (a) a joint RGB model, trained under Robometer's token-causal mask and reading the mean of the two views' progress tokens; (b) the same model with patch and Q/K rays; (c) the causal mask replaced by synchronous block attention, reading the last view's progress token; and (d) the self-readout mask added to (c), which is the full \method{}. Within a task, all runs share the data, camera pairs, LoRA configuration, and seed. Run (a) is trained for 12k steps and (b)--(d) for 24k; the step from (b) to (c) changes both the mask and the readout, whereas (c) and (d) differ only in the self-readout mask.

Adding rays under the causal mask (b) lowers far-azimuth MAE on three tasks but raises it on PushCube; switching to block attention with last-view readout (c) lowers it on three tasks but raises it on StackCube, by about half. Only the self-readout mask (d) makes the gain consistent: it lowers MAE on all four tasks and raises $\tau_a$ on three, leaving PushCube within 0.001. Relative to the joint RGB model (a), the full configuration lowers MAE by 26--39\% and raises $\tau_a$ on every task.

\begin{table}[t]
\centering\footnotesize
\caption{Ray-path ablation on PickCube, all arms under synchronous block attention with the self-readout mask and a matched 24k budget, differing only in the rays given to the backbone. Test MAE$\downarrow$ and $\tau_a\uparrow$ on the far-azimuth group and averaged over the four camera groups.}
\label{tab:ablation_rays}
\setlength{\tabcolsep}{4pt}
\begin{tabular}{lcccc}\toprule
& \multicolumn{2}{c}{Far azimuth} & \multicolumn{2}{c}{Mean of groups}\\
\cmidrule(lr){2-3}\cmidrule(lr){4-5}
Rays & MAE & $\tau_a$ & MAE & $\tau_a$\\\midrule
None (RGB) & .0371 & .891 & .0392 & .889\\
Patch & .0403 & .864 & .0410 & .867\\
Q/K & .0354 & \textbf{.892} & .0376 & \textbf{.890}\\
Patch + Q/K & \textbf{.0335} & .876 & \textbf{.0346} & .877\\
\bottomrule
\end{tabular}
\end{table}

Table~\ref{tab:ablation_rays} separates the two ray paths on PickCube with the attention configuration and the budget of the full model held fixed, so the arms differ only in what the ray encoder supplies. The lowest MAE is obtained when both paths are present: patch rays alone are worse than no rays on both metrics, Q/K rays alone lower far-azimuth MAE from 0.0371 to 0.0354, and the two together reach 0.0335, about 10\% below the arm without rays. Ordering does not follow error: $\tau_a$ is 0.891 without rays and 0.892 with Q/K rays alone, against 0.876 for the full model, a gap twice the step-to-step spread of validation $\tau_a$ over the second half of training (standard deviation 0.007--0.008). Checkpoints are selected by validation MAE and never by $\tau_a$, and on validation the selected full model has the higher $\tau_a$ of the two (0.897 versus 0.874), so we do not read the test ordering as a cost of the rays.

This series also bounds what the first table can attribute to geometry. Its reference (a) is a causal 12k run, whereas an RGB arm trained under block attention with the self-readout mask reaches far-azimuth MAE 0.0371 against 0.0494 for (a), so most of the improvement from (a) to (d) comes from the attention configuration rather than from the rays. With one camera the rays account for the improvement instead, lowering MAE by 14--27\% (Sec.~IV-B), whereas here they add 10--14\% on top of the attention configuration.

\section{Conclusion and Limitations}
We presented \method, a framework for adapting a pretrained robotic reward model to a task using the geometry of the cameras that observe it. Its contributions are threefold. First, token-aligned Pl\"ucker rays condition both the visual features and the attention queries and keys during low-rank adaptation, so one adapted model accepts any single calibrated view; on PickCube it generalizes better than RGB adaptation to unseen viewpoints, reducing MAE under a changed field of view by approximately 21\%. Second, synchronous block attention lets synchronized views attend to one another inside the pretrained decoder without new parameters, so several cameras are processed as one sequence; across four simulated tasks, joint prediction lowers MAE by 41--69\% relative to late fusion of single-view RGB models, and the ablations show that the self-readout mask lowers MAE on all four tasks. Third, on three real tasks observed by a fixed and a wrist-mounted camera, joint prediction has the lowest error and the best ordering on every task, reducing mean MAE by approximately 21\% relative to averaged RGB fine-tuning and following reversed, frozen, and interrupted clips more closely.

The current evidence supports a specific deployment strategy: adapt an existing robotic reward model to task-specific progress prediction, supply its observation geometry, and add a complementary camera when a single view hides relevant state. Processing the two views jointly with their camera geometry, which together define \method{}, yields a consistent improvement over late fusion of single-view models.

Several limits remain. All training comparisons are single runs without multi-seed intervals. Camera calibration is required~\cite{tsailenz}, and sensitivity to real calibration errors is not yet established. Most supervised progress results concern labeled trajectories, and the real-robot demonstrations carry time-linear labels without failures, so neither establishes robust failure detection~\cite{badbehavior}. Finally, on the real robot we evaluate the adapted model only offline, as a reward model scoring recorded demonstrations; we have not yet trained a policy with its progress estimates as rewards, so whether more accurate progress estimates translate into faster or more reliable reinforcement learning remains an open question.

\end{document}